\documentclass{article}
\usepackage{spconf,amsmath,graphicx,hyperref}
\usepackage[inline]{enumitem} 
\usepackage{xspace}
\usepackage{amssymb}
\usepackage{multirow}
\usepackage{booktabs}
\usepackage{xcolor}

\newcommand{\methodname}{PDG\xspace}
\newcommand{\cy}{city A\xspace}
\newcommand{\cz}{city B\xspace}

\title{Physics-informed Diffusion Generation for Geomagnetic \\ Map Interpolation}

\name{
    \begin{tabular}{@{}c@{}}
        Wenda Li$^{1}$ \qquad 
        Tongya Zheng$^{2}$ \qquad 
        Kaixuan Chen$^{1}$ \qquad 
        Shunyu Liu$^{1}$ \qquad 
        Haoze Jiang$^{1}$ \\
        Yunzhi Hao$^{2}$ \qquad
        Rui Miao$^{3}$ \qquad 
        Zujie Ren$^{3\dagger}$ \qquad 
        Mingli Song$^{1}$ \qquad 
        Hang Shi$^{3}$ \qquad 
        Gang Chen$^{1}$
    \end{tabular}
    \thanks{$^\dagger$ Corresponding author:~Zujie Ren~(renzju@zju.edu.cn)}
}

\address{
    $^{1}$ State Key Laboratory of Blockchain and Data Security, Zhejiang University \\
    $^{2}$ High-Performance Intelligent Computing Research Center for Ultra-Large Scale Graph Data, \\ School of Computer and Computing Science, Hangzhou City University \\
    $^{3} $ Zhejiang Lab \\
  }
\begin{document}
\maketitle
\begin{abstract}
Geomagnetic map interpolation aims to infer unobserved geomagnetic data at spatial points, yielding critical applications in navigation and resource exploration.
However, existing methods for scattered data interpolation are not specifically designed for geomagnetic maps, which inevitably leads to suboptimal performance due to detection noise and the laws of physics.
Therefore, we propose a \textbf{P}hysics-informed \textbf{D}iffusion \textbf{G}eneration framework~(\methodname) to interpolate incomplete geomagnetic maps.
First, we design a physics-informed mask strategy to guide the diffusion generation process based on a local receptive field, effectively eliminating noise interference.
Second, we impose a physics-informed constraint on the diffusion generation results following the kriging principle of geomagnetic maps, ensuring strict adherence to the laws of physics.
Extensive experiments and in-depth analyses on four real-world datasets demonstrate the superiority and effectiveness of each component of \methodname.
\end{abstract}
\section{Introduction}
\label{sec:intro}
Geomagnetic map interpolation~\cite{lee2015geomagnetic, aleshin2022geomagnetic} aims to infer unobserved magnetic data in space and is widely applied in navigation~\cite{zhao2021summary}, resource exploration~\cite{irfan2024integrative}, and precise positioning~\cite{goldenberg2006geomagnetic}. 
In practical measurements, environmental factors often cause the measurement trajectory to exhibit a chain-like distribution, and the collected data typically contain noise.
Traditional methods~\cite{cressie1990origins,lu2008adaptive,buhmann2000radial} are based on the principle of local consistency~\cite{alldredge1980local, lesur2006introducing} in geomagnetic data, where the geomagnetic field varies smoothly and continuously across neighboring regions, estimating values at unobserved points using nearby observations and assigning spatial weights according to explicitly defined functional relationships.
However, when applied to large-scale datasets that often contain noise, these methods typically face challenges in model performance.

Recently, deep learning-based methods have emerged to capture latent correlations between observed and unobserved points for scattered data interpolation, which shares similar data formats with geomagnetic data. Neural Processes predict the distribution of target points from context data using conditional encoding~\cite{garnelo2018conditional}, attention mechanisms~\cite{lu2008adaptive}, or bootstrapping~\cite{lee2020bootstrapping} to estimate uncertainty. 
NIERT~\cite{ding2024niert} adopts a pre-trained Transformer on synthetic functions to improve interpolation and generalization. 
HINT~\cite{ding2023accurate} hierarchically leverages observed-point residuals to iteratively refine interpolation with lightweight modules.
Despite the effective spatial correlation modeling, the aforementioned methods are not particularly designed for geomagnetic map interpolation.
Specifically, two challenges remain in scattered data interpolation in geomagnetic data.
First, these methods typically consider clean data without accounting for noise, resulting in significant disturbances in geomagnetic data due to noise. 
Second, neural modules often disrupt physical smoothness and continuity because of their strong nonlinearity, thereby violating the laws of physics in the geomagnetic map.

To address the above challenges, we propose a \textbf{P}hysics-informed \textbf{D}iffusion \textbf{G}eneration for geomagnetic map interpolation~(\methodname).
First, we introduce a \textbf{geomagnetic diffusion model} that interpolates geomagnetic data through a step-wise iterative generation process, effectively suppressing noise in the observations.
To further reduce noise, we design a \textbf{physics-informed mask} strategy that dynamically adjusts the local receptive field during the diffusion process, guiding data generation with physical principles. 
To ensure adherence of the diffusion-generated results to physical principles, we incorporate a \textbf{physics-informed constraint} guided by the Kriging approach.
Extensive experiments on four real-world geomagnetic datasets show that \methodname reduces the interpolation error by up to 80\%.
Visualization analysis further illustrates its superiority in local areas, and comprehensive ablation studies verify the effectiveness of each component.

\section{Preliminary}
\textbf{Geomagnetic Map Interpolation.}
For a spatial coordinate $m^i = (\mathrm{lon}_i, \mathrm{lat}_i)$, the corresponding geomagnetic field intensity is $x^i \in \mathbb{R}^d$, measured in nanoteslas~(nT). Geomagnetic map interpolation aims to predict the magnetic field intensity $x^{ta}$ at a target location $m^{ta}$, given the coordinates $m^o = \{m^1, \dots, m^n\}$ and magnetic field intensity measurements $x^o = \{x^1, \dots, x^n\}$ of multiple observed points:
\begin{equation}
    x^{ta} = f(m^{o}, x^{o}, m^{ta}),
\end{equation}
where $f(\cdot)$ denotes the interpolation function that estimates the field at an unknown location based on the observed data.

\noindent
\textbf{Diffusion Models.}
Diffusion models~\cite{croitoru2023diffusion, cao2024survey} are probabilistic generative models rooted in principles of non-equilibrium thermodynamics and stochastic differential equations. 
A canonical example is the Denoising Diffusion Probabilistic Model (DDPM)~\cite{ho2020denoising}, which comprises a forward process for noise injection and a reverse process for generating data from Gaussian noise.
During the forward process, an initial input ${x}_0 \sim q({x}_0)$ is gradually corrupted into a Gaussian noise vector through $t$ steps that can be described as a Markov chain:
\begin{equation}
    q({x}_t | {x}_{t-1}) = \mathcal{N}\left(\sqrt{1 - \beta_t} {x}_{t-1}, \beta_t I\right), 1 \le t \le T,
\end{equation}
where $\beta_t \in [0,1]$ represents the noise level at step $t$. Alternatively, the distribution of ${x}_t$ conditioned directly on ${x}_0$ can be written as 
$q({x}_t | {x}_0) = \mathcal{N}\left({x}_t; \sqrt{\bar{\alpha}_t} {x}_0, (1 - \bar{\alpha}_t) I\right),$
where $\bar{\alpha}_t = \prod_{s=1}^t \alpha_s$ and $\alpha_t = 1 - \beta_t$. Thus, ${x}_t$ can be simply obtained as
\begin{equation}
\label{addNoise}
    {x}_t = \sqrt{\bar{\alpha}_t} {x}_0 + \sqrt{1 - \bar{\alpha}_t} \epsilon,
\end{equation}
where $\epsilon$ is standard Gaussian noise.
During the reverse process, a neural network model $\epsilon_\theta$ is used to learn the denoising distribution
$p_\theta(x_{t-1}\mid x_t) = \mathcal{N}\big(x_{t-1}; \mu_\theta(x_t, t), \sigma_\theta(x_t, t)\big),$
where the variance $\sigma_\theta({x}_t, t)$ is often fixed to  $\sigma^2 I$, and the mean $\mu_\theta({x}_t, t)$ is computed as $\mu_\theta({x}_t, t) = \frac{1}{\sqrt{\alpha_t}} {x}_t - \frac{1 - \alpha_t}{\sqrt{1 - \bar{\alpha}_t} \sqrt{\alpha_t}} \epsilon_\theta({x}_t, t)$.
The training objective is to minimize the following loss:
\begin{equation}
    \mathcal{L}_\epsilon = \mathbb{E}_{t, {x}_0, \epsilon} \left\| \epsilon - \epsilon_\theta({x}_t, t) \right\|_2^2.
\end{equation}

\section{Method}
\begin{figure}[t]
  \centering
  \includegraphics[width=\linewidth]{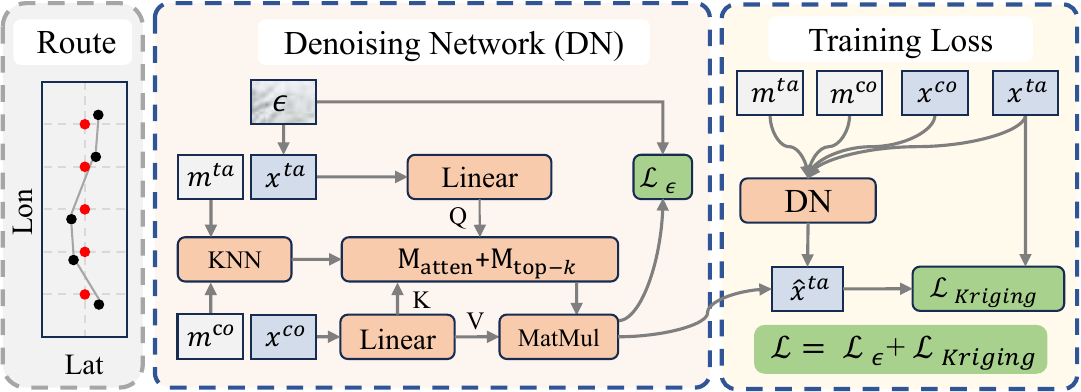}
  \caption{The overall framework of \methodname.}
  \label{graph:main}
\end{figure}

In this section, we introduce \methodname, a novel physics-informed diffusion generation framework for geomagnetic map interpolation. The architecture of our method is shown in Figure~\ref{graph:main}.

\subsection{Physics-Informed Geomagnetic Diffusion}
Noise in measurement data, caused by environmental factors, affects the interpolation of geomagnetic maps. To reduce the impact of noise, we employ a diffusion model to reconstruct missing data through a step-by-step denoising process. Based on the physical principle of local consistency, we design a physics-informed mask strategy that dynamically adjusts the local neighborhood range at each diffusion step to suppress noise interference further while generating smooth data.
A conditional diffusion model~\cite{zhang2023adding} is adopted to predict missing geomagnetic data, leveraging known observations as spatial cues to guide the prediction of missing values more accurately compared with an unconditional diffusion model that generates samples solely from noise.

Given the observed data pairs $(m^{o}, x^{o})$ and the target data pairs $(m^{ta}, x^{ta})$.
The objective of the denoising model is to estimate the noise added to the magnetic field intensities $x^{ta}$ at the target locations $m^{ta}$.
At the specific diffusion step $t$, Gaussian noise $\boldsymbol{\epsilon} \sim \mathcal{N}(0, \mathrm{I})$ is added to the target magnetic field intensities to generate the noisy inputs as Equation~\ref{addNoise}:
\begin{equation}
    \label{eq:addnoise}
     x_t^{ta}=\sqrt{\Bar{\alpha}_t}x_0^{ta}+\sqrt{1-\Bar{\alpha}_t}\epsilon,
\end{equation}
where ${x}^{ta}_0$ represents the true magnetic field intensities at $m^{ta}$.
The observed data pairs $(m^{o}, x^{o})$, which are used as conditions $(m^{co}, x^{co})$, are fed into the denoising model along with the noised target pairs $(m^{ta}, x^{ta}_t)$.
We first apply linear projections to the conditional pair $(m^{co}, x^{co})$ and target pair $(m^{ta}, x^{ta})$ to obtain their latent representations: $h_t = \text{Linear}(m_t, x_t) + {p}^t$, where ${p}^t = \mathrm{FFN}(\mathrm{Emb}(t))$ and 
$\mathrm{Emb}(t)=\bigl(\sin(10^{\tfrac{4i}{w-1}}t)\bigr)_{i=0}^{w-1}\ \Vert\ \bigl(\cos(10^{\tfrac{4i}{w-1}}t)\bigr)_{i=0}^{w-1}$.
Here, $\text{Linear}(\cdot)$ denotes a learnable linear projection layer, $\mathrm{FFN}(\cdot)$ refers to the Feedforward Neural Network, ${p}^t$ is the diffusion step embedding at step $t$, $\text{Emb}(t)$ is a $d$-dimensional vector for step encoding, and $w=d/2$.

In the physics-informed mask, for each target point, the receptive field is restricted to the top-$k$ geographically closest conditional points, thereby reducing noise from points outside the receptive field and generating smooth data.
Specifically, $\mathrm{M}_{{\mathrm{top}-{k}}} = \mathbf{1}(m^{co} \in \text{KNN}\left(m^{ta}\right)).$
As the denoising process progresses, the model generates increasingly accurate data, so the receptive field of each target point is set as a function of the diffusion step, i.e., $k = K(t)$. In the early stages, a larger $K$ is used to incorporate more conditional points for coarse estimation, while a smaller $K$ is applied in later stages to capture fine-grained local patterns.
Specifically, $K(t) = K_{\min} + \frac{t}{T} \times \left(K_{\max} - K_{\min}\right),$ 
where $t$ is the current diffusion timestep, $T$ is the total number of diffusion steps, $K_{\min}$ is the minimum neighborhood size used in the final stages of denoising, and $K_{\max}$ is the maximum neighborhood size used in the initial stages.

A cross-attention mechanism is utilized to model the spatial dependencies between target points and available conditional points.
First, we obtain the query, key, and value vectors through linear projections where
${Q} = {h}^{ta}_t {W}^Q, \quad {K} = {h}^{co}_t {W}^K, $ and$ \quad {V} = {h}^{co}_t {W}^V.$
\begin{equation}
    {o_t^{ta}} = \operatorname{Softmax} \left( \frac{{Q} {K}^\top}{\sqrt{D}} \odot \mathrm{M}_{{\mathrm{top}-{k}}} \right) {V},
\end{equation}
where $\operatorname{Softmax}(\cdot)$ normalizes the input scores into a probability distribution, ${W}^Q$, ${W}^K$, and ${W}^V$ are learnable weight matrices for the query, key, and value projections, $D$ is the dimensionality of the query and key vectors, used for scaling.
Then, the predicted noise $\hat{\epsilon}^{ta}_{t}$ is obtained via a residual connection and an MLP, i.e., $\hat{\epsilon}^{ta}_{t} = \operatorname{MLP}\big(h^{ta}_t + o^{ta}_t\big)$.
At diffusion step $t$, the denoised magnetic field intensities $\hat{{x}}^{ta}_0$ is recovered using the standard DDPM reverse formulation:
\begin{equation}
    \label{eq:denoise_xta}
    \hat{{x}}^{ta}_0 = \frac{ {x}^{ta}_t - \sqrt{1 - \Bar{\alpha}_t} \cdot \hat{\epsilon}^{ta}_t }{ \sqrt{\Bar{\alpha}_t} }.
\end{equation}

\subsection{Kriging-Guided Physics-Informed Constraint}
To enhance the physical consistency of predictions, we introduce a kriging-guided physics-informed loss. This loss draws on the principle of kriging to model spatial autocorrelation~\cite{cressie1990origins}, using the similarity of nearby points to constrain the neural network outputs to reflect the spatial variations observed in the real geomagnetic field.
For any pair of locations $i$ and $j$, we define the empirical variogram based on the ground truth and predicted values as $\gamma_{\text{true}}(r^{ij}) = \frac{1}{2} (x^{ta, i} - x^{ta, j})^2$ and $\gamma_{\text{pred}}(r^{ij}) = \frac{1}{2} (\hat{x}^{ta, i} - \hat{x}^{ta, j})^2$, where $r_{ij} = | m^{ta, i} - m^{ta, j} |$ denotes the Euclidean distance between the target locations $m^{ta, i}$ and $m^{ta, j}$.

To capture local spatial structures while maintaining computational efficiency, for each target location $i$, we select its $t$ nearest neighbors, denoted as $\mathcal{N}_t(i)$. The kriging-guided physics-informed loss is then formulated as:
\begin{equation}
    \label{eq:loss_kriging}
    \mathcal{L}_{\text{Kriging}} = \frac{1}{|m^{ta}|} \sum_{i=1}^{|m^{ta}|} \frac{1}{t} \sum_{j \in \mathcal{N}_t(i)} \left( \gamma_{\text{pred}}(r^{ij}) - \gamma_{\text{true}}(r^{ij}) \right)^2,
\end{equation}
where $|m^{ta}|$ is the number of target locations.

\subsection{Training Process}

During the training process, the denoising loss at diffusion step $t$ is defined as the mean squared error between the true noise and the predicted noise.
During training, the denoising loss at diffusion step $t$ is defined as the mean squared error between the true noise and the predicted noise, i.e., $\mathcal{L}_\epsilon = \mathbb{E}_{{x}^{ta}_0, t, \boldsymbol{\epsilon} \sim \mathcal{N}(0, \mathrm{I})}\big[\|\epsilon - \epsilon_\theta(x_t, t)\|^2\big]$.
Together with the kriging-guided physics-informed loss. $\mathcal{L}_{\text{Kriging}}$, the total loss $\mathcal{L}$ is
\begin{equation}
    \label{eq:loss_all}
    \mathcal{L} = \mathcal{L}_\epsilon + \lambda \mathcal{L}_{\text{Kriging}}.
\end{equation}
where $\lambda$ is a weighted coefficient.

\noindent
\section{Experiments}
\noindent
\subsection{Experiment Settings}
\noindent
\textbf{Datasets.}
We collected geomagnetic data along UAV flight paths in the \cy\footnotemark{}\addtocounter{footnote}{-1} and \cz\footnotemark{} regions.
\footnotetext{To preserve confidentiality, the identities of the two cities analyzed in this study are anonymized and represented by abbreviations.}
For \cy, datasets A-InX, A-InZ, and A-OutZ are constructed from $in$-cabin components X and Z and the $out$-of-cabin Z component, respectively. For \cz, dataset B-InT is based on the $in$-cabin total field intensity T. All datasets are split into training, validation, and test sets in an 8:1:1 ratio.
All experiments were performed using an NVIDIA A800 GPU.

\noindent
\textbf{Baselines.}
To evaluate the interpolation accuracy, we compared \methodname with existing representative neural network-based interpolation methods, including Conditional Neural Processes~(CNP)~\cite{garnelo2018conditional}, Attentive Neural Processes~(ANP)~\cite{kim2019attentive}, Bootstrapping Attentive Neural Processes~(BANP)~\cite{lee2020bootstrapping}, NIERT~\cite{ding2024niert}, TFR-Transformer~\cite{chen2021machine}, and HINT~\cite{ding2023accurate}.

\noindent
\subsection{Results}
\textbf{Quantitative Results.}
Table~\ref{table_all} shows that \methodname reduces interpolation error by 80\% on average across four real-world datasets, while other deep-learning methods perform much worse with errors several times larger.
Although HINT achieves comparable performance on the A-InX dataset, it underperforms on the remaining datasets and suffers from out-of-memory (OOM) issues, underscoring the necessity of our framework for both accurate and efficient geomagnetic interpolation.

\begin{table}
    \centering
    \small
    \caption{Overall performance comparison, with the best result on each dataset highlighted in  \textbf{bold}. ``OOM'' stands for ``Out of Memory''. }
    \label{table_all}
    \resizebox{\linewidth}{!}{
        \begin{tabular}{cccccccc|c}
            \toprule
            Dataset     & Metric     & CNP     & BANP  & ANP      & TFR              & Niert    & HINT  & \methodname \\
            \midrule
            \multirow{4}{*}{A-InX}  & RMSE & 2.511   & 2.363 & 2.696    & 2.135           & 1.320    & 0.560 & \textbf{0.552}       \\
                                     & MAE  & 2.511   & 2.064 & 2.362    & 1.866           & 1.085    & 0.449 & \textbf{0.422}       \\
                                     & MAPE & 0.783   & 2.064 & 0.947    & 0.636           & 0.357    & 0.449 & \textbf{0.144}       \\
                                     & MSE  & 7.942   & 7.187 & 10.089   & 6.647           & 1.965    & 0.341 & \textbf{0.332}       \\
            \midrule
            \multirow{4}{*}{A-InZ}  & RMSE & 2.038   & 1.931 & 5.657    & 1.595           & 2.650    & 0.886 & \textbf{0.719}       \\
                                     & MAE  & 1.734   & 1.592 & 4.636    & 1.388           & 2.155    & 0.665 & \textbf{0.556}       \\
                                     & MAPE & 0.696   & 1.592 & 1.900    & 0.411           & 0.603    & 0.170 & \textbf{0.132}       \\
                                     & MSE  & 4.592   & 4.184 & 47.590   & 3.080           & 7.689    & 0.917 & \textbf{0.540}       \\
            \midrule
            \multirow{4}{*}{B-InT}  & RMSE & 23.355  & OOM   & 43.260   & 44.637          & 31.811   & OOM   & \textbf{1.092}       \\
                                     & MAE  & 18.781  & OOM   & 35.492   & 34.931          & 24.613   & OOM   & \textbf{0.511}       \\
                                     & MAPE & 0.573   & OOM   & 0.840    & 1.161           & 0.964    & OOM   & \textbf{0.013}       \\
                                     & MSE  & 635.520 & OOM   & 1982.100 & 2162.436        & 1099.566 & OOM   & \textbf{7.997}       \\
            \midrule
            \multirow{4}{*}{A-OutZ} & RMSE & 2.189   & 2.525 & 3.508    & 2.217           & 1.340    & 0.564 & \textbf{0.288}       \\
                                     & MAE  & 1.863   & 2.200 & 3.069    & 1.952           & 1.083    & 0.470 & \textbf{0.227}       \\
                                     & MAPE & 0.554   & 0.661 & 1.188    & 0.617           & 0.238    & 0.173 & \textbf{0.064}       \\
                                     & MSE  & 5.686   & 8.360 & 16.642   & 6.131           & 2.215    & 0.337 & \textbf{0.120}       \\
            \bottomrule
        \end{tabular}
    }
\end{table}

\begin{table}
    \centering
    \small
    \caption{Ablation study on \methodname. ``PIM'' and ``PIC'' stand for physics-informed mask and kriging-guided physics-informed loss, respectively.}
    \label{table_ablation}
    \resizebox{\linewidth}{!}{\begin{tabular}{lllllllllllll}
        \toprule
        \multicolumn{1}{c}{\multirow{2}{*}{Methods}} & \multicolumn{2}{c}{A-InX} & \multicolumn{2}{c}{A-InZ} & \multicolumn{2}{c}{B-InT} & \multicolumn{2}{c}{A-OutZ} \\
        \cmidrule(lr){2-3}\cmidrule(lr){4-5}\cmidrule(lr){6-7}\cmidrule(lr){8-9}  
        \multicolumn{1}{c}{} & \multicolumn{1}{c}{RMSE} & \multicolumn{1}{c}{MAPE} & \multicolumn{1}{c}{RMSE} & \multicolumn{1}{c}{MAPE} & \multicolumn{1}{c}{RMSE} & \multicolumn{1}{c}{MAPE} & \multicolumn{1}{c}{RMSE} & \multicolumn{1}{c}{MAPE} \\
        \midrule
        w/o PIM          & 0.609 & 0.145 & 0.896 & 0.140 & 2.019 & 0.032 & 0.348 & 0.071 \\
        w/o PIC & 0.631 & 0.155 & 0.913 & 0.146 & 2.020 & 0.025 & 0.337 & 0.064 \\
        \methodname      & \textbf{0.552} & \textbf{0.144} & \textbf{0.719} & \textbf{0.132} & \textbf{1.092} & \textbf{0.013} & \textbf{0.288} & \textbf{0.064} \\
        \bottomrule                 
    \end{tabular}}
\end{table}

\begin{figure}[h]
  \centering
  \includegraphics[width=\linewidth]{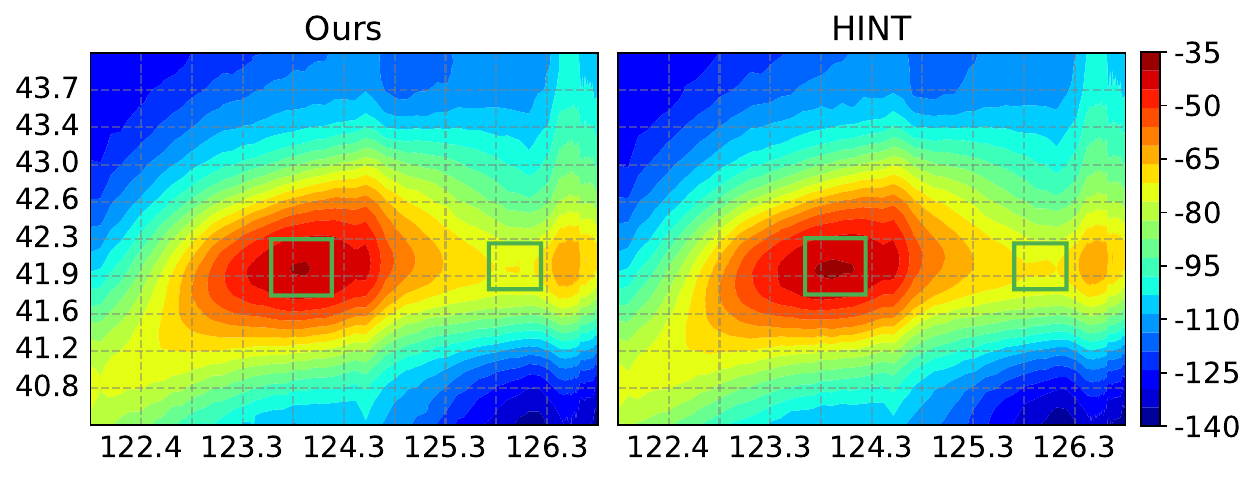}
  \caption{High-precision geomagnetic map of dataset A-OutZ. The x-axis denotes longitude, and the y-axis denotes latitude.}
  \label{fig_vis}
\end{figure}

\noindent
\textbf{Qualitative Results.}
Figure~\ref{fig_vis} shows that our method and the baseline HINT both capture the large-scale geomagnetic distribution well, exhibiting consistent patterns in major anomaly regions such as the central high-value zone. However, in local areas, our method performs better: the contour lines are smoother with more natural transitions, effectively suppressing noise. In contrast, HINT shows fluctuations and spikes, with abrupt high or low values near certain points, indicating higher sensitivity to outliers. Overall, our method is more robust and produces more realistic and reliable geomagnetic maps.

\noindent
\textbf{Ablation Study of Different Components.}
To more thoroughly evaluate \methodname, we conducted ablation studies by removing key components. As shown in Table~\ref{table_ablation}, removing either the physics-informed mask or the kriging-guided physics-informed loss degrades performance, indicating that both noise handling and adherence to local physical consistency are crucial for improving interpolation accuracy in geomagnetic map interpolation tasks.

\noindent
\textbf{Parameter Sensitivity Study on Sampling Steps.}
We conducted a parameter sensitivity experiment to investigate the impact of sampling steps. Six sampling steps were selected: 5, 10, 20, 30, 40, and 50. As shown in Table~\ref{table_sampling_steps}, setting the sampling steps within the range of 5–20 achieves a good balance between performance and efficiency. This indicates that excessively large sampling steps do not necessarily improve accuracy but instead increase computational overhead, whereas moderate sampling steps can provide a better trade-off between accuracy and efficiency.

\noindent
\textbf{Parameter Sensitivity Study on $K_{\max}$ and $K_{\min}$.}
We analyzed the impact of different values of $K_{\max}$ and $K_{\min}$ on the interpolation performance. Table~\ref{table_sampling_L} shows that $K_{\max}$ has a significant effect on the completion results. This is because, during the initial steps, the neighborhood size is large and the geomagnetic data exhibits strong fluctuations. Selecting an appropriate range is crucial for capturing the overall spatial patterns and ensuring accurate interpolation. In contrast, $K_{\min}$ has a relatively minor impact on the final results. As the number of steps decreases, the neighborhood size becomes smaller and the geomagnetic fluctuations are limited, meaning the model primarily serves to refine and consolidate the interpolation results rather than improve them substantially.

\noindent
\section{Conclusion}
In this paper, we propose \methodname, a physics-informed diffusion generation framework for geomagnetic map interpolation. By integrating a geomagnetic diffusion model, a physics-informed mask, and physical constraints, our method effectively suppresses noise and enforces physical consistency. Experiments on four real-world datasets show up to 80\% error reduction and superior performance in irregular regions, while ablation and visualization studies further demonstrate the effectiveness and advantages of each component.

\begin{table}[]
    \centering
    \small
    \caption{Parameter sensitivity study on sampling steps.}
    \label{table_sampling_steps}
    \resizebox{\linewidth}{!}{\begin{tabular}{lllllllll}
        \toprule
        \multicolumn{1}{c}{\multirow{2}{*}{step}} & \multicolumn{2}{c}{A-InX} & \multicolumn{2}{c}{A-InZ} & \multicolumn{2}{c}{B-InT} & \multicolumn{2}{c}{A-OutZ} \\
        \cmidrule(lr){2-3}\cmidrule(lr){4-5}\cmidrule(lr){6-7}\cmidrule(lr){8-9}  
        \multicolumn{1}{c}{} & \multicolumn{1}{c}{RMSE} & \multicolumn{1}{c}{MAPE} & \multicolumn{1}{c}{RMSE} & \multicolumn{1}{c}{MAPE} & \multicolumn{1}{c}{RMSE} & \multicolumn{1}{c}{MAPE} & \multicolumn{1}{c}{RMSE} & \multicolumn{1}{c}{MAPE} \\
        \midrule
        5  & 0.492 & 0.143 & 0.636 & 0.122 & 0.998 & 0.014 & 0.366 & 0.074 \\
        10 & 0.552 & 0.144 & 0.719 & 0.132 & 1.092 & 0.013 & 0.282 & 0.064 \\
        20 & 0.555 & 0.150 & 0.690 & 0.138 & 1.044 & 0.014 & 0.282 & 0.068 \\
        30 & 0.571 & 0.150 & 0.725 & 0.140 & 1.013 & 0.013 & 0.285 & 0.072 \\
        40 & 0.575 & 0.146 & 0.733 & 0.144 & 1.094 & 0.014 & 0.287 & 0.080 \\
        50 & 0.577 & 0.153 & 0.748 & 0.147 & 1.033 & 0.014 & 0.290 & 0.066 \\
        \bottomrule                 
    \end{tabular}}
\end{table}

\begin{table}[]
    \centering
    \small
    \caption{Parameter sensitivity study on $K_{\max}$ and $K_{\min}$.}
    \label{table_sampling_L}
    \resizebox{\linewidth}{!}{
        \begin{tabular}{c|c|cccccccc}
            \toprule
            \multicolumn{1}{c|}{\multirow{2}{*}{${K_{\max}}$}} & \multicolumn{1}{c|}{\multirow{2}{*}{${K_{\min}}$}} & \multicolumn{2}{c}{A-InX} & \multicolumn{2}{c}{A-InZ} & \multicolumn{2}{c}{B-InT} & \multicolumn{2}{c}{A-OutZ} \\
            \cmidrule(lr){3-4}\cmidrule(lr){5-6}\cmidrule(lr){7-8}\cmidrule(lr){9-10}  
              &  & \multicolumn{1}{c}{RMSE} & \multicolumn{1}{c}{MAPE} & \multicolumn{1}{c}{RMSE} & \multicolumn{1}{c}{MAPE} & \multicolumn{1}{c}{RMSE} & \multicolumn{1}{c}{MAPE} & \multicolumn{1}{c}{RMSE} & \multicolumn{1}{c}{MAPE} \\
            \midrule
            \multirow{2}{*}{1500} & 32 & 0.561 & 0.141 & 0.864 & 0.142 & 1.901 & 0.016 & 0.307 & 0.064 \\
                                  & 64 & 0.562 & 0.142 & 0.866 & 0.143 & 1.945 & 0.016 & 0.308 & 0.064 \\
            \midrule
            \multirow{2}{*}{1000} & 32 & \textbf{0.552} & \textbf{0.141} & 0.724 & 0.137 & \textbf{1.092} & \textbf{0.013} & \textbf{0.288} & \textbf{0.062} \\
                                  & 64 & 0.552 & 0.141 & 0.727 & 0.138 & 1.132 & 0.013 & 0.289 & 0.063 \\
            \midrule
            \multirow{2}{*}{500}  & 32 & 0.554 & 0.143 & \textbf{0.719} & \textbf{0.132} & 1.129 & 0.018 & 0.302 & 0.059 \\
                                  & 64 & 0.555 & 0.143 & 0.719 & 0.132 & 1.171 & 0.019 & 0.303 & 0.060 \\
            \bottomrule                 
        \end{tabular}
    }
\end{table}

\noindent
\textbf{Relation to Prior Work.} 
The work presented here focuses on the development of a geomagnetic data interpolation algorithm that accounts for both noise suppression and compliance with physical principles. 
In contrast, the work by Shizhe and Dongbo~\cite{ding2023accurate} reduces interpolation errors through a hierarchical residual optimization method.
While the present study is related to recent scattered data interpolation methods~\cite{lee2020bootstrapping,ding2024niert,ding2023accurate}, it is specifically designed for geomagnetic data and effectively exploits geomagnetic characteristics that were not addressed in earlier studies.

\noindent
\section{Acknowledgements}
This work was supported by the National Natural Science Foundation of China (Grant No. 62506330), the Zhejiang Provincial Natural Science Foundation of China (Grant No. LQN26F020007), Zhejiang Province High-Level Talents Special Support Program "Leading Talent of Technological Innovation of Ten-Thousands Talents Program" (No.2022R52046), the Fundamental Research Funds for the Central Universities (2021FZZX001-23), the advanced computing resources provided by the Super computing Center of Hangzhou City University, the Key R\&D Program of Zhejiang (2024C01036).

\bibliographystyle{IEEEbib}
\bibliography{refs}

\end{document}